\documentclass[journal]{IEEEtran}

\usepackage{amsmath,amsfonts}
\usepackage{algorithm}
\usepackage{algpseudocode}
\usepackage{array}
\usepackage[]{subfig}
\usepackage{textcomp}
\usepackage{stfloats}
\usepackage{url}
\usepackage{verbatim}

\usepackage{graphicx}
\usepackage{cite}
\usepackage{hyperref}
\usepackage{bm}
\usepackage{makecell}
\usepackage{footnote}
\usepackage{float}
\usepackage{subfig}
\usepackage{bbding}
\usepackage{ragged2e}
\usepackage{amsthm}
\usepackage{dsfont}
\usepackage{lipsum} 
\usepackage{booktabs}
\usepackage{multirow}
\usepackage{xcolor}
\usepackage{makecell}
\usepackage[normalem]{ulem}
\useunder{\uline}{\ul}{}

\usepackage{booktabs}      
\usepackage[table]{xcolor} 

\usepackage{enumitem} 
\usepackage{pifont}

\makeatletter

\newcommand{\Rmnum}[1]{\expandafter\@slowromancap\romannumeral #1@}
\makeatother

\theoremstyle{definition}

\usepackage{cuted}

\usepackage[switch]{lineno}

\begin{document}

\title{When Can We Trust Deep Neural Networks? Towards Reliable Industrial Deployment with an Interpretability Guide}

\author{Hang-Cheng Dong, Yuhao Jiang, Yibo Jiao, Lu Zou, Kai Zheng, Bingguo Liu, Dong Ye and Guodong Liu


\thanks{\textit{(Corresponding authors: Bingguo Liu)}}

}


\maketitle

\begin{abstract}
The deployment of AI systems in safety-critical domains, such as industrial defect inspection, autonomous driving, and medical diagnosis, is severely hampered by their lack of reliability. A single undetected erroneous prediction can lead to catastrophic outcomes. Unfortunately, there is often no alternative but to place trust in the outputs of a trained AI system, which operates without an internal safeguard to flag unreliable predictions, even in cases of high accuracy. We propose a post-hoc explanation-based indicator to detect false negatives in binary defect detection networks. To our knowledge, this is the first method to proactively identify potentially erroneous network outputs. Our core idea leverages the difference between class-specific discriminative heatmaps and class-agnostic ones. We compute the difference in their intersection over union (IoU) as a reliability score. An adversarial enhancement method is further introduced to amplify this disparity. Evaluations on two industrial defect detection benchmarks show our method effectively identifies false negatives. With adversarial enhancement, it achieves 100\% recall, albeit with a trade-off for true negatives. Our work thus advocates for a new and trustworthy deployment paradigm: data-model-explanation-output, moving beyond conventional end-to-end systems to provide critical support for reliable AI in real-world applications.

\end{abstract}


\section{Introduction}
\label{sec:introduction}
\IEEEPARstart{C}{urrently}, deep learning methods have made significant advancements in both image~\cite{SAM, ravi2024sam2} and language~\cite{llmcomprehensive} domains. In particular, the remarkable success in the field of computer vision has enabled its application across various industrial scenarios, such as defect detection~\cite{ameri2024systematic, ma2024surface}, autonomous driving~\cite{reda2024path}, and medical diagnosis~\cite{bougourzi2025recent}. However, the inherent flaws of deep learning severely hinder its deployment in critical domains. The current deep learning paradigm lacks reliability, and a single error may lead to unacceptable catastrophic consequences. Even by increasing the volume of training data, impressive accuracy rates can be achieved, but this does not alter its black-box nature~\cite{fan}. Without any intrinsic mechanism to proactively flag potential errors of the trained model, users can only blindly trust the model's judgments.

The deep learning paradigm can be summarized as a data-model-output pipeline. An intuitive approach is to assess the credibility of the output based on the magnitude of the model's prediction. In addition, uncertainty estimation~\cite{uncertainty} methods have emerged that attempt to quantify model confidence using statistical sampling techniques~\cite{solmcdrop}. Both types of methods essentially rely directly on the model's output, with their core focus being on what the model does not know. Thus, they are often applied to out-of-distribution (OOD) detection~\cite{yang2024generalized}. Crucially, these methods often remain ineffective against predictions that are "in-distribution yet confidently wrong," and such failure modes represent a critical vulnerability in safety-critical applications.

Can we determine when the network makes an error? Our core hypothesis is that the reliability of a prediction should be judged by the model's internal reasoning process. Therefore, we turn our attention to the interpretability of deep learning. Post-hoc explanation methods such as Grad-CAM~\cite{gradcam} can reveal the primary contributing regions that support the current model decision. Depending on their relevance to the output class, explanations can be categorized as discriminative or non-discriminative. This discrepancy inspires us to explore whether we can assess the reliability of the network's decision by comparing different explanations for the same sample.

Therefore, we re-examine post-hoc explanation methods and validate them in defect detection scenarios. To simplify the discussion, we refer to the two outputs of the defect detection problem as the background class and the defect class, corresponding to negative and positive samples, respectively. We find that different types of visual explanation methods produce discrepancies, which may stem from fundamental differences in the internal features they capture.
We propose to quantify this discrepancy as the IoU difference between explanatory heatmaps, termed $\Delta$-IoU. As an indicator of model safety performance, $\Delta$-IoU can effectively identify false negative samples, but it can also lead to misclassification of true negative samples. Thus, we refer to samples with anomalous $\Delta$-IoU as suspicious samples. To enable $\Delta$-IoU to detect all false negative samples, we further propose an adversarial enhancement method. By performing adversarial attacks on the samples, subtle defect features are accentuated.

To the best of our knowledge, this is the first proposed criterion for determining when a model's output may be erroneous. The overall workflow demonstrates the irreplaceable role of interpretability research. Therefore, we argue that the classical data-model-output paradigm can be revised into a new paradigm: data-model-interpretation-output, which may lay the foundation for future AI applications in mission-critical domains. In summary, the contributions of this paper can be summarized as follows:

\begin{itemize}
    
\item We propose a training-free method requiring no architectural modifications to identify suspicious samples, thereby preventing the potential severe harm caused by false negatives.

\item The proposed adversarial enhancement method achieves 100\% recall in defect detection tasks, providing new insights for the application of deep learning methods in mission-critical domains. 

\item We introduce interpretability as an integral component of deep learning model decision-making, integrating it into the decision-making process to form a new data-model-interpretation-output paradigm. Building upon this, we propose the novel task of suspicious sample detection, laying the foundation for applications in safety-critical domains.

\end{itemize}

\section{Related work}

\label{sec:related}

\textbf{Uncertainty Estimation.} Uncertainty estimation serves as a primary methodology for assessing the reliability of deep learning models~\cite{uncertainty}. The sources of uncertainty in network outputs can be categorized into aleatoric and epistemic types~\cite{Aleatoric, hora1996aleatory}. Techniques such as MC dropout~\cite{JMLRDropout,solmcdrop} and deep ensembles typically quantify predictive confidence by measuring variations in model outputs or parameters. In essence, uncertainty estimation excels at identifying what a model does not know, making it particularly effective for tasks like out-of-distribution detection~\cite{yang2024generalized}. 

In contrast, this work focuses on instances where models are confidently wrong, which is a scenario we term suspicious samples. We define the task of identifying such potentially erroneous model outputs as suspicious sample detection (SSD). Confidence-based approaches like uncertainty estimation and our proposed suspicious sample detection are fundamentally distinct in both their objectives and methodologies.

\textbf{Explainable AI.} Explainable AI (XAI) techniques aim to reveal the internal decision-making processes of deep learning models~\cite{fan}. Based on how explanations are generated, these techniques can be categorized into built-in interpretable models~\cite{chattopadhyay2022interpretablebydesign, CoDANet} and post-hoc interpretation methods. Built-in interpretable models, such as B-CosNet~\cite{BCos}, achieve layer-wise feature explanations by eliminating bias terms. However, as most models lack this inherent property~\cite{Rudin2019}, post-hoc interpretation methods are employed to uncover the internal decision logic of such models. A classic approach for feature interpretation is feature visualization~\cite{2015Understanding, Dconv}. For instance, gradient-based explanations~\cite{grad1, grad2} can reflect the contribution of input pixels to the final decision. Class activation mapping (CAM)~\cite{36cam}, on the other hand, enables analysis of the last convolutional features by assigning weights to channels based on gradient statistics, thereby generating visual heatmaps. Building upon this, numerous variants~\cite{gradcam, gradcamplus, scam, zhang2021group} have been developed that assign feature weights differently. For example, layer-wise relevance propagation (LRP)~\cite{lrp} and its variants~\cite{lrp1, lrp2} distribute relevance scores backward through the network. 

In summary, feature visualization methods provide valuable insights into the internal decision mechanisms of networks. Depending on whether the explanation method is related to the output category, it can be classified as discriminative or non-discriminative. Specifically, Grad-CAM~\cite{gradcam} falls under discriminative methods, while FullGrad~\cite{fullgrad} is categorized as a non-discriminative approach.

\textbf{Industrial Defect Detection.} Industrial defect detection is a classic application for computer vision~\cite{ren2022state}. Early defect detection methods relied on hand-crafted feature extractors, following classical machine vision approaches~\cite{GOLNABI2007630}. Deep learning has now revolutionized machine vision methodologies, leading to a proliferation of deep learning-based research in industrial defect detection~\cite{ameri2024systematic, ma2024surface}. Currently, data-driven paradigms based on convolutional neural networks~\cite{zheng2024md} and transformers~\cite{yu2024railway} have become the state-of-the-art benchmark.

However, inherent limitations in data-driven approaches prevent AI from being directly deployed in critical production processes. A model may achieve high overall accuracy while remaining vulnerable to unexpected high-confidence failures. Catastrophic false negative samples must be eliminated in safety-critical applications.

\section{Methodology}
\label{sec:formatting}
\subsection{Post-hoc explanation}

\begin{figure*}[htbp]
    \centering
    \includegraphics[width=1\linewidth]{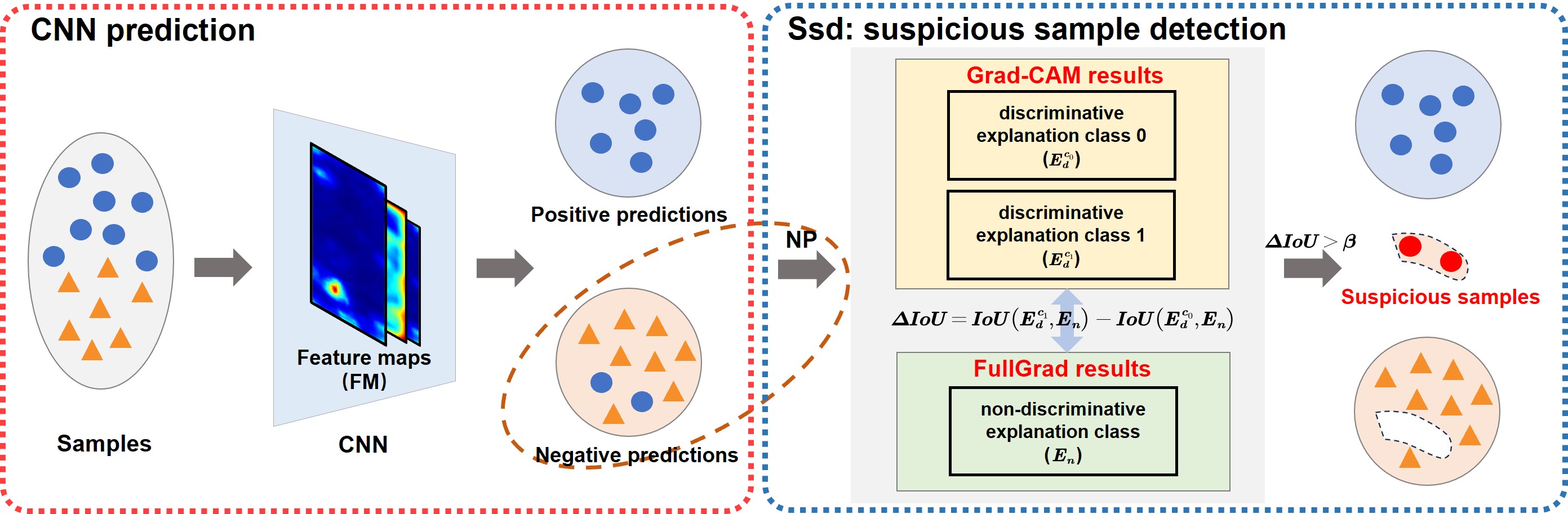}
    \caption{The proposed framework for suspicious sample detection.}
    \label{fig:framework}
\end{figure*}

\begin{figure*}[ht]
    \centering
    \includegraphics[width=1\linewidth]{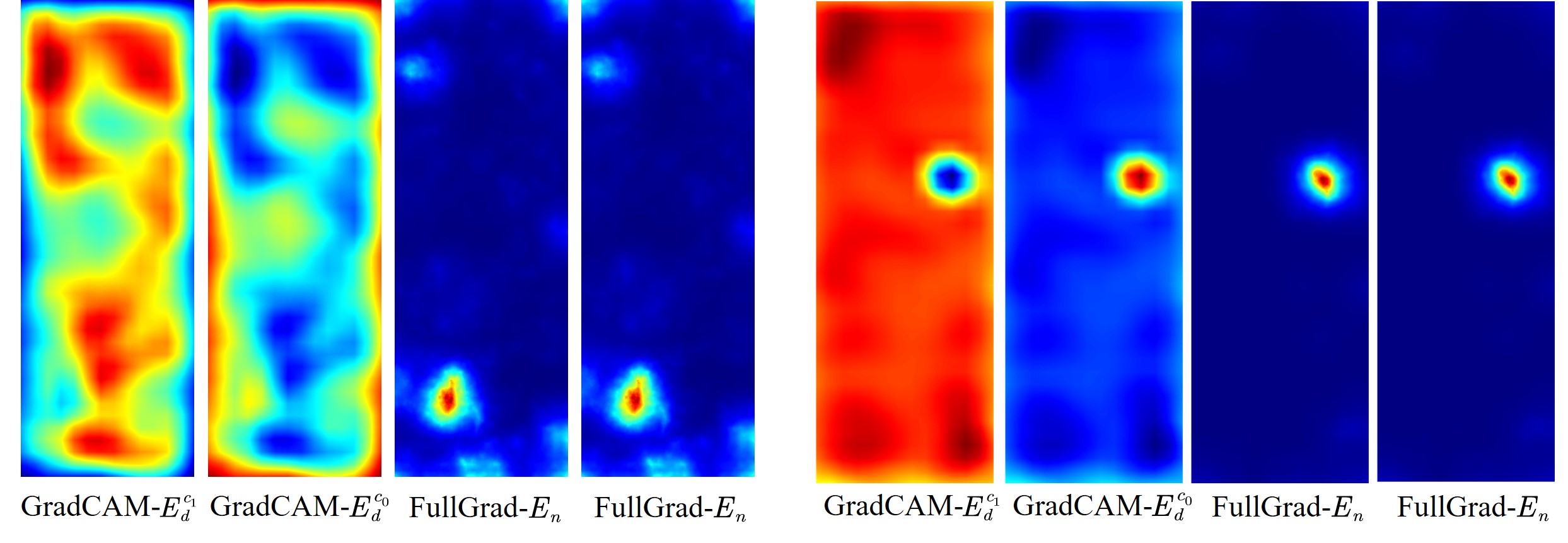}
    \caption{Visual explanations for two selected samples (one True Negative and one False Negative), featuring four heatmaps generated by two distinct methods for both output classes. The left panel depicts an FN case, showing from left to right: Grad-CAM for positive class, Grad-CAM for negative class, FullGrad for positive class, and FullGrad for negative class. The right panel presents a TN case following the same arrangement. Notably, the two Grad-CAM heatmaps for each sample exhibit complementary characteristics, whereas the FullGrad visualizations remain nearly identical across classes. All explanations were generated from a convolutional neural network trained on the Kolektor SDD2 dataset.}
    \label{fig:222}
\end{figure*}

Post-hoc explanations can aid in understanding the internal decision-making processes of neural networks. We employ two classical post-hoc explanation methods, Grad-CAM and FullGrad, whose fundamental definitions are provided below.

Mathematically, consider a pre-trained classification model $f$ comprising $L$ convolutional layers. For an input image $\mathbf{I}$ with category $c$, the pre-softmax logit is denoted as $y^c$, where $\mathbf{\theta}$ represents the model parameters. Let $\mathbf{A}^{l}\in \mathbb{R}^{W_l \times H_l \times C_l}$ be the activation tensor from the $l$-th layer, with $W_l$, $H_l$ and $C_l$ are spatial dimensions and channel count of $l$-th feature map respectively, and $C_l$ is the number of the channels in $l$-th convolutional layer, respectively. 


\textbf{Grad-CAM.} Firstly, Grad-CAM~\cite{gradcam} computes the global average of channel-wise weight $w_{kl}^c$ by 
\begin{equation}
    w_{kl}^c = \frac{1}{Z_l} \sum_i \sum_j \frac{\partial y^c}{\partial A_{ij}^{kl}},
    \label{eq2}
\end{equation}
where $Z_l = W_l \times H_l$. Then, the saliency map for layer $l$ is obtained by linearly combining the corresponding feature maps as

\begin{equation}
    M_{\text{Grad-CAM}}^{cl} = \text{ReLU} (\sum_k w_{kl}^c \cdot \mathbf{A}^{kl}),
    \label{eq3}
\end{equation}
where a ReLU function suppresses negative values of the saliency map. 

\textbf{FullGrad.} Grad-CAM solely considers the features from the final convolutional layer, whereas FullGrad~\cite{fullgrad} incorporates the influence of bias terms by aggregating gradient information from all layers, including the input layer. Its formulation is as follows:

\begin{equation}
    M^{c}_{\text{FullGrad}} = \frac{\partial y^c}{\partial \mathbf{x}} \otimes \mathbf{x} + \sum_{l=1}^{L}\sum_{k=1}^{C_l} \Psi(\frac{\partial y^c}{\partial \mathbf{b}^{kl}} \otimes \mathbf{b}^{kl}),
\end{equation}
where $\mathbf{x}$ denotes the input image, $\mathbf{b^{kl}}$ is the bias of $k$-th neuron in layer $l$, and $\Psi(\cdot)$ performs post-processing function. It is worth noting that to maintain alignment between resolution and input dimensions, Grad-CAM also employs interpolation methods, such as bilinear upscaling, to resize its heat maps to match the input image size.
\begin{figure*}[ht]
    \centering
    \includegraphics[width=1\linewidth]{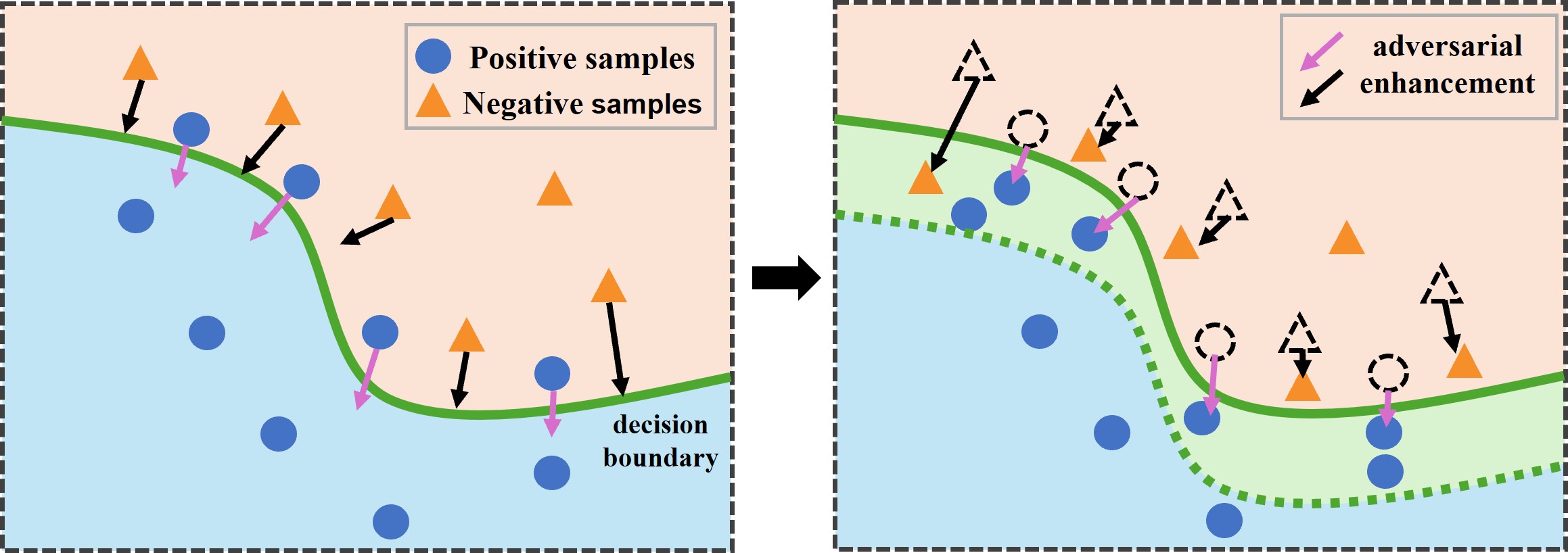}
    \caption{Schematic diagram of the proposed adversarial enhancement method.}
    \label{fig:333}
\end{figure*}
Grad-CAM constitutes a discriminative explanation method, indicating that its generated heatmaps are class-specific. In contrast, FullGrad represents the class-agnostic approach, meaning it captures features relevant to all non-background categories. As summarized in Table \ref{tab1}, we list the properties of several common explanation methods.

A remaining open question concerns what determines the discriminative nature of an explanation method. Our hypothesis posits that the features extracted by the explanation method fundamentally govern its discriminative property. Grad-CAM leverages feature maps from the final layer, which directly serve as classification criteria, thus exhibiting strong discriminative characteristics. Conversely, FullGrad utilizes gradient information that amalgamates multi-layer features, wherein earlier convolutional layers have not yet learned discriminative representations.

\subsection{Suspicious sample detection}

Can the internal decision-making of neural networks be used to determine when the network makes an error? Our research starts from this question. To facilitate reading, we first present some simple but key concepts. For a binary classification task, given an input $x$, the trained model outputs $y$. Our goal is to determine whether y is a true negative (TN) or a false negative (FN) when the output $y$ is negative.

As shown in Figure \ref{fig:222}, we generate distinct explanations for different sample types (TN and FN). The discriminative method, Grad-CAM, is denoted as $e_d$, while the non-discriminative method is denoted as $e_n$. Specifically, the discriminative method can generate two types of explanations based on the output classes, which we denote as $e_d^{c_0}$ and $e_d^{c_1}$, where the label "0" represents the negative (background) class and "1" represents the positive (defect) class.


According to Figure \ref{fig:222}, it can be clearly observed that since the FullGrad method always tends to find all non-background features in the model, for the FN sample, both FullGrad and Grad-CAM methods will locate the defect regions, thus they are similar. For the TN sample, FullGrad and Grad-CAM differ significantly because at this point, defect features are scarce, and mainly background features are present. 

Formally, we can use a simple intersection over union (IoU) to quantify the difference between $e_n$ and $e_d$. We set a threshold for the generated heatmaps and then binarize them, commonly using the Otsu method to generate an adaptive threshold. Denoting the obtained images as $E_n$ and $E_d$, the aforementioned difference can be mathematically expressed as

\begin{equation}
    \Delta IoU = IoU(E_d^{c_1},E_n)-IoU(E_d^{c_0},E_n).
    \label{eq4}
\end{equation}

Based on this observation, we propose a simple yet efficient indicator $\Delta IoU$ to alert about the significant risk of potential missed detections by the network. Figure \ref{fig:framework} shows the complete flowchart of our method.

\textbf{Adversarial enhancement.} Due to differences in model training, some false negative samples may exhibit insufficiently prominent features. To ensure all false negative samples can be detected, we propose an adversarial enhancement method~\cite{goodfellow2014explaining}, as shown in Figure \ref{fig:333}. For an input sample $x$, we shift the feature toward the positive class $y_{c_1}$, thereby amplifying the defect features and making false negative samples easier to detect. Specifically, the iterative formula is as follows:
\begin{equation}
    x^t = x^{t-1}+\alpha \frac{\partial y_{c_1}}{\partial x^{t-1}},
    \label{eq5}
\end{equation}
where $t$ denotes the iteration number and $\alpha$ represents the step size.

After performing several iterations of adversarial enhancement on the whole dataset, Equation (\ref{eq4}) can be computed to enable suspicious sample detection.

\begin{table}
  \caption{Properties of four classic visual explanation methods~\cite{grad1,lrp,gradcam,fullgrad}. Based on the source of features and weights, they can be divided into gradient-based, activation-based, and attribution-based categories. "Class specific" indicates whether the method is discriminative. Except for Grad-CAM, all other methods back-propagate to the input domain.}
  \label{tab1}
  \centering
  \begin{tabular}{@{}lcccr@{}}
    \toprule
     & Int. Grad & LRP & Grad-CAM & FullGrad  \\
    \midrule
    Gradients & \checkmark &              &  \checkmark& \checkmark\\
    Attribution &           & \checkmark && \\
    Feature Maps &          &            &   \checkmark&\\
    Class Specific &        &  & \checkmark &\\
    Input Domain & \checkmark & \checkmark & & \checkmark \\
    \bottomrule
  \end{tabular}
\end{table}


\section{Experimental results}

\subsection{Experimental setup}


We evaluate our method on two classic industrial defect detection benchmarks. Both datasets originate from surface defects of electronic commutators. The first dataset, Kolektor SDD~\cite{ksdd}, contains 399 samples in total, including 52 defective samples. We approximately split the data into training and test sets in a 2:1 ratio, where the test set contains 16 defective samples and 105 defect-free samples. Similarly, the Kolektor SDD2~\cite{ksdd2} dataset is divided into two parts: the training set contains 2085 defect-free samples and 246 defective samples, while the test set contains 894 defect-free samples and 110 defective samples.

We employ several commonly used metrics in classification problems to comprehensively evaluate the performance of the suspicious sample detection task, specifically the counts of true negatives (TN) and false negatives (FN), as well as recall and accuracy. The formula for recall is given as follows:



\begin{equation}
    Recall = \frac{TP}{TP+FN}.
\end{equation}


\subsection{Main results}
As shown in Table \ref{tab:KSDD}, we present the performance of our proposed suspicious sample detection method on the Kolektor SDD dataset. For simplicity, we abbreviate our method as $\Delta IoU$. We trained a base VGG16 model, denoted as "origin", and then applied the proposed $\Delta IoU$ method for detection. We set the threshold $\beta$ of $\Delta IoU$ to 0.2, meaning $\Delta IoU$ of the sample exceeding 0.2 is considered suspicious. It should be noted that $\beta$ can be set within a relatively broad range.
For comparison, we used confidence-based thresholding as our baseline, with the confidence threshold set at 0.95. Evidently, confidence cannot serve as an effective means to assess whether the model has made an error, which fundamentally differs from our approach. Meanwhile, it can be observed that our method accurately identifies existing suspicious samples. After excluding suspicious samples, the model achieves 100\% recall.

\begin{table}
  \caption{Suspicious sample detection results on the Kolektor SDD dataset.}
  \label{tab:KSDD}
  \centering
  \begin{tabular}{@{}lcccc@{}}
    \toprule
    Method & TN & FN $\downarrow$ & Recall $\uparrow$ & Acc(\%) $\uparrow$ \\
    \midrule
    Origin    & 105 & 1 & 93.4 & 99.2 \\
    Confidence & 103 & 1 & 93.4 & 99.2 \\
    \rowcolor{gray!30} $\Delta IoU$ & 84 & 0 & 100.0 & 100.0 \\
    \bottomrule
  \end{tabular}
\end{table}

As shown in Table \ref{tab:ssdvgg}, we report the suspicious sample detection results on the Kolektor SDD2 dataset. The experimental setup remains consistent with that in Table \ref{tab:KSDD}. The confidence-based method only identified 2 out of 19 potential risks, with merely marginal improvements in both recall and precision. This evidence further demonstrates that model outputs should not be blindly trusted, as numerous potential risks are output by the network with high confidence. In contrast, our proposed method successfully identified 10 out of 19 FNs, achieving an increase of 8.3\% in recall and 0.9\% in accuracy compared to the original model. While our method currently has the limitation of misclassifying 64 TN samples as suspicious, it demonstrates overall beneficial performance.

\begin{table}[ht]
  \caption{Suspicious sample detection results on the Kolektor SDD2 dataset.}
  \label{tab:ssdvgg}
  \centering
  \begin{tabular}{@{}l|ccccccr@{}}
    \toprule
 & TN  & FN $\downarrow$ & Recall $\uparrow$ &  Acc(\%) $\uparrow$ \\
    \midrule
    Origin & 887 & 19 & 82.7& 97.4 \\
    Confidence &  874 &  17 & 84.3 & 97.6         \\
\rowcolor{gray!30}    $\Delta IoU$ & 823 & 9 &91.0& 98.3          \\

    \bottomrule
  \end{tabular}
\end{table}

Table \ref{tab:ssdvggcbam} reports the suspicious sample detection results with another model on the Kolektor SDD2 dataset. We trained a VGG16 model with CBAM~\cite{woo2018cbam} attention to investigate how suspicious sample detection performance varies across different model architectures. Our method $\Delta IoU$ successfully identified 10 out of 14 potential risks, leaving only 4 false negative samples undetected, achieving a 96.0\% recall rate. This demonstrates the reliability of our approach. Furthermore, comparing Tables \ref{tab:ssdvgg} and \ref{tab:ssdvggcbam}, we observe that models with better feature extraction capabilities tend to achieve higher recall rates, though potentially at the cost of misclassifying more true negative samples.

\begin{table}[ht]
  \caption{Suspicious sample detection results on the Kolektor SDD2 dataset. The model employed is a VGG16 architecture integrated with the CBAM mechanism.}
  \label{tab:ssdvggcbam}
  \centering
  \begin{tabular}{@{}l|ccccccr@{}}
    \toprule
 & TN  & FN $\downarrow$ & Recall $\uparrow$ &  Acc(\%) $\uparrow$ \\
    \midrule
    Origin(CBAM) & 881 & 14 & 87.3& 97.3 \\
    
\rowcolor{gray!30}    $\Delta IoU$ & 763 & 4 & 96.0 & 98.1          \\

    \bottomrule
  \end{tabular}
\end{table}


\subsection{The effect of adversarial enhancement for SSD}

Now, we discuss the feasibility of achieving 100\% recall to meet the detection requirements in safety-critical domains. Table \ref{tab:adv} reports the specific impact of the adversarial enhancement strategy on suspicious sample detection, referred to $\Delta IoU_{adv}$. Using the same baseline models as in Tables 3 and 4, we implemented adversarial enhancement with a step size of 0.01. Given the substantial impact of adversarial enhancement on the overall model, we applied only mild enhancement by setting the number of iterations to 2. 

On the original VGG16 baseline, our method $\Delta IoU_{adv}$ successfully identified 17 out of 19 potential risks, significantly increasing recall by 15.5\%. For the CBAM-enhanced model, our approach successfully excluded all risks, achieving 100\% recall. However, this comes at the cost of a substantial increase in misclassified true negative samples, which decreased from 881 to 250.

\begin{table}
  \caption{Suspicious sample detection results with adversarial enhancement on the Kolektor SDD2 dataset.}
  \label{tab:adv}
  \centering
  \begin{tabular}{@{}l|ccccccr@{}}
    \toprule
 & TN  & FN $\downarrow$ & Recall $\uparrow$ &  Acc(\%) $\uparrow$ \\
    \midrule
    Origin & 887 & 19 & 82.7& 97.4 \\
    \rowcolor{gray!30}    $\Delta IoU_{adv}$(Origin) & 419 & 2 & 98.2 & 98.3 \\
    Origin(CBAM) & 881 & 14 & 87.3& 97.3  \\

\rowcolor{gray!30}    $\Delta IoU_{adv}$(CBAM) &  250 &  0 & 100.0 &   96.4       \\

    \bottomrule
  \end{tabular}
\end{table}


\section{Discussion and limitations}

Our experimental results validate the initial observation that integrating interpretability into the final decision-making process substantially mitigates potential security risks. However, our current framework represents merely an initial step toward this goal. The experimental outcomes reveal two persistent open challenges requiring further investigation. 

The first challenge concerns the origin of discriminative features. Specifically, whether they derive from shallow or deep network layers as hypothesized, and whether more effective feature extraction methodologies exist. The second challenge lies in the relatively high rate of false positives, where numerous true negative samples are incorrectly flagged as suspicious. Through careful analysis of these misclassified samples, we identified that many exhibit visual characteristics bearing resemblance to genuine defects. This observation suggests that current limitations stem from two primary sources: the model's feature representation capacity and the insufficient incorporation of domain-specific knowledge.  Future work will address these aspects through two main directions: extending the framework to multi-category detection scenarios, and developing methods to extract more precise knowledge representations from within the model architecture itself.

\section{Conclusion}

This study analyzes the important role of interpretability in deep learning decision-making. The proposed new paradigm of "data-model-interpretation-output" provides a novel perspective for identifying potential erroneous predictions. First, we propose a simple yet efficient suspicious sample detection method based on the discrepancy between discriminative and non-discriminative explanations. To the best of our knowledge, our method reveals that the model's internal decision-making can be used as a basis for judging model reliability in the first time. Second, the developed adversarial enhancement strategy effectively improves the salience of defect features. Finally, experiments on industrial defect detection benchmarks demonstrate that this method can significantly reduce security risks, providing important safeguards for applications in critical domains.

\bibliographystyle{IEEEtran}
\bibliography{ref.bib}

\end{document}